\documentclass{article}
\usepackage{ijcai26}
\usepackage{times}
\usepackage{soul}
\usepackage{url}
\usepackage[hidelinks]{hyperref}
\usepackage[utf8]{inputenc}
\usepackage[small]{caption}
\usepackage{graphicx}
\usepackage{amsmath}
\usepackage{amsthm}
\usepackage{amssymb}
\usepackage{booktabs}
\usepackage{algorithm}
\usepackage{algorithmic}
\usepackage[switch]{lineno}

\usepackage{booktabs}
\usepackage{tikz}
\usetikzlibrary{shapes.geometric, arrows.meta, positioning, fit, calc, backgrounds, shadows}

\definecolor{encoderblue}{RGB}{230, 240, 255}
\definecolor{decoderorange}{RGB}{255, 240, 230}
\definecolor{contextgreen}{RGB}{230, 255, 235}
\definecolor{darkblue}{RGB}{0, 50, 100}

\title{Context-Aware Concept Distillation for Trustworthy Flood Prediction}

\author{
    Eli Levinkopf$^1$ \and
    Efrat Morin$^2$ \And
    Claudia V. Goldman$^3$\\
    \affiliations
    $^1$School of Computer Science and Engineering, The Hebrew University of Jerusalem, Israel\\
    $^2$Institute of Earth Sciences, The Hebrew University of Jerusalem, Israel\\
    $^3$Hebrew University Business School, The Hebrew University of Jerusalem, Israel\\
    \emails
    \{eli.levinkopf, efrat.morin, claudia.goldman\}@mail.huji.ac.il
}

\newcommand{\updated}[1]{#1}

\begin{document}

\maketitle

\begin{abstract}
Effective flood risk management relies on accurate forecasting, yet the "black box" nature of state-of-the-art Deep Learning models creates a barrier to trust and accountability in high-stakes public safety decisions. While existing Explainable AI (XAI) methods offer local attributions, they fail to provide the verifiable, operationally meaningful causal narratives required by disaster response authorities. To address this societal challenge, we propose \textbf{Context-Aware Concept Distillation (CACD)}, a framework developed in collaboration with domain experts to distill opaque LSTMs into interpretable, hydrology-aware surrogate models. We introduce an unsupervised pipeline to discover a "Hydrological Language" and a Residual Hypernetwork that dynamically modulates these concepts based on static basin characteristics. Evaluated on 5,203 basins globally, our model achieves high fidelity (Median NSE 0.70), significantly outperforming black-box baselines (e.g., Multi Layer Perceptrons) on unseen future data. By demonstrating that human-interpretable concepts are sufficient to reconstruct flood dynamics, this work balances AI accuracy with the transparency required for responsible environmental decision-making.
\end{abstract}

\section{Introduction}

AI systems for flood forecasting have reached unprecedented scale and accuracy, exemplified by recent global initiatives for ungauged watersheds \cite{nearing2024global}. Such systems are outstanding in making remote data accessible and facilitating early warnings, effectively democratizing access to predictive information. However, authorities and emergency responders managing flooded areas need to trust these predictions and understand the outputs beyond a raw number. Systems that lack transparency face significant hurdles in adoption for decision support and public communication \cite{dikshit2024artificial}. Therefore, explainable AI (XAI) capabilities are essential to transform these powerful computational systems into trustworthy and interpretable tools \cite{visave2025transparency,vu2025blackbox}.

Currently, public authorities routinely accompany flood forecasts with causal narratives, such as "intense short-duration rainfall" or "prolonged precipitation over saturated soils", to enhance credibility. While recent studies have employed interpretable machine-learning approaches to identify flood-generating mechanisms \cite{jiang2022river}, such efforts often remain qualitative and do not demonstrate whether the extracted explanations are operationally meaningful. This limits their utility for accountability. To bridge this gap, we introduce \textbf{Context-Aware Concept Distillation (CACD)}, a framework that represents flood dynamics through interpretable concepts derived from SHAP-based clustering. Unlike traditional methods, we explicitly validate usability by reconstructing flood magnitude using an encoder–decoder architecture. By demonstrating that these concepts are not only interpretable but also sufficient to re-simulate flood behavior, we move beyond descriptive explainability toward verifiable, decision-relevant understanding.

Our framework relies on a fundamental insight: hydrological reasoning is intrinsically low-dimensional but context-dependent. A complex flood event can be described by a small vocabulary of dynamic drivers (e.g., precipitation), but the impact of these drivers on streamflow is strictly modulated by the static physical attributes of the basin (e.g., slope, soil permeability). We thus train a surrogate model governed by a Residual Hypernetwork, decoupling the prediction into global hydrological laws and local physiographic adaptations.

\textbf{Societal Problem:} We address the "Black Box Trust Gap" in flood risk management (UN SDG 13). Decision-makers cannot rely on opaque AI predictions for life-critical early warnings without verifiable justification.
\textbf{Stakeholder Roles:} The hydrological domain expert co-authoring this paper provided the data and the original model for streamflow prediction. She was actively involved in the XAI development, contributing domain expertise to assess the explanations according to their use in real-world hydrological practice. The work is further informed by her sustained, close collaboration with national-level water management and watershed authorities involved in flood prediction and risk mitigation, ensuring that the proposed methods address operational needs and support trustworthy deployment in public safety contexts.
\textbf{Impact Path:} The proposed methodology bridges machine learning forecasting systems and trustworthy decision-making, enabling authorities to audit AI predictions against established hydrological knowledge before issuing public alerts. The main impact lies in transforming powerful predictive models into reliable AI decision-support tools.

\noindent \textbf{Our Contributions:} (1) \textbf{Unsupervised Concept Language:} We propose a high-fidelity feature engineering pipeline using log-temporal windowing to extract semantic concepts from raw attribution scores, eliminating the need for manual concept labeling. (2) \textbf{Hydrology-Aware Architecture:} We introduce a Context-Aware Decoder based on a Residual Hypernetwork. This design imposes a structural inductive bias that models streamflow as the product of temporal forcing and spatial sensitivity. (3) \textbf{Superior Generalization via Structural Bias:} We demonstrate that our interpretable surrogate significantly outperforms a standard black-box decoder (Deep MLP) on unseen future data (Median Fidelity NSE 0.70 vs. 0.60). This suggests that the structural constraints imposed by our architecture act as a powerful regularizer, capturing the teacher's reasoning rules better than a generic non-linear model.
\updated{\noindent \textbf{Code \& Reproducibility:} The complete codebase, pre-trained surrogate models, and the extracted concept dataset are open-sourced at \url{https://github.com/eli-levinkopf/cacd-flood}.}

\section{Related Work}

\subsection{Deep Learning in Hydrology}
The hydrological modeling community has seen a paradigm shift toward Deep Learning, driven by the availability of large-scale datasets such as CAMELS and Caravan \cite{kratzert2023caravan}. In particular, Long Short-Term Memory (LSTM) networks have emerged as the state-of-the-art for rainfall-runoff modeling, consistently outperforming traditional process-based models in regional benchmarks \cite{kratzert2019benchmarking} and operational forecasting frameworks \cite{nevo2022flood}. \updated{While our distillation framework is architecture-agnostic, we employ an LSTM teacher because recent literature demonstrates they still outperform Transformers in hydrological simulations \cite{hess-26-3377-2022,LIU2024131389}.}

Kratzert et al.~\shortcite{kratzert2019benchmarking} integrated static basin attributes to modulate LSTM processing. While our decoder also uses static features for conditioning, our goals differ fundamentally: their approach learns an opaque, high-dimensional embedding ($\mathbb{R}^{256}$) optimized solely for performance. In contrast, we enforce a low-dimensional semantic bottleneck ($\mathbb{R}^6$) designed explicitly for interpretability.

\subsection{Explainability in Earth Systems}
The deployment of AI for high-stakes climate and weather challenges requires more than just accuracy; it demands interpretability to calibrate user trust and facilitate scientific discovery \cite{mamalakis2022xai}. This is particularly critical for modeling extreme events such as flash floods, where understanding the physical controls—such as runoff-contributing areas \cite{rinat2018controls} and radar rainfall patterns \cite{morin2009flash}—is essential for risk assessment.

However, reviews of XAI applications in hydrology \cite{bramm2025review} indicate that the field relies predominantly on post-hoc feature attribution. As surveyed by Rojat et al.~\shortcite{rojat2021explainable}, these methods fall into two main categories. First, Backpropagation-based methods (e.g., Saliency Maps \cite{simonyan2014deep}, Grad-CAM \cite{selvaraju2017grad}) compute gradients of the output with respect to input time steps. Second, Perturbation-based methods measure prediction changes when inputs are masked (e.g., SHAP \cite{lundberg2017unified}, LIME \cite{ribeiro2016should}). These methods suffer from a lack of semantic interpretability and global consistency. A key limitation of LIME is that it implicitly trains thousands of disjoint local linear models, offering no coherent global logic. Similarly, raw SHAP values provide local attribution but fail to distinguish between different hydrological mechanisms (e.g., "Snowmelt" vs. "Flash Flood") across the entire dataset. Our work bridges this gap by aggregating atomic SHAP values into higher-level semantic concepts.

\subsection{Concept-Based Interpretability}
To move beyond low-level features (pixels or time steps), recent work has proposed analyzing models in terms of human-understandable concepts. Kim et al.~\shortcite{kim2018interpretability} introduced Testing with Concept Activation Vectors (TCAV), demonstrating that high-dimensional internal states of neural networks can be mapped to linear concept vectors.

This line of reasoning led to Concept Bottleneck Models (CBMs) \cite{koh2020concept}, which enforce a strict separation where the model first predicts concepts (e.g., ``Wing Color'') and then predicts the target via a linear function. However, standard CBMs suffer from two limitations in the hydrological context:
\textbf{Supervision Bottleneck:} They require dense concept labels (e.g., expert annotation for every time step), which do not exist at the scale of the Caravan dataset (spanning millions of daily time steps across thousands of basins). \updated{Beyond this vast scale, it is cognitively infeasible for a human expert to manually process 65 interacting sequential parameters to label even a single event. Lacking pre-existing labels, unsupervised clustering combined with post-hoc expert validation is the only practical approach.}
\textbf{Static Mappings:} They typically assume a fixed relationship between concepts and targets. However, in hydrology, the relationship is context-dependent: the concept ``Heavy Rain'' leads to different prediction outcomes when referring to data in arid bare-soil versus humid forested basins.

Our \textit{Context-Aware Concept Distillation} framework overcomes these limitations by discovering concepts in an unsupervised manner (via SHAP clustering) and utilizing a Residual Hypernetwork to dynamically adapt the concept-to-prediction mapping based on static basin attributes.

% ----------------------------------------------------------------------------------------------------------------

\section{Methodology}

We propose the \textit{Context-Aware Concept Distillation} (CACD) framework, designed to interpret the reasoning of black-box sequence-based models in the physical sciences (demonstrated here on hydrological LSTMs). 

Our approach is built on the hypothesis that the reasoning process of complex models is intrinsically low-dimensional but context-dependent. That is, we assume that the high-dimensional internal representations of a trained LSTM can be compressed into a small vocabulary of dynamic concepts (e.g., "Snowmelt", "Flash Flood"), provided that the mapping from these concepts to streamflow predictions is modulated by the static physical attributes of the specific basin.

% --- FIGURE 1 ---
\begin{figure*}[t]
    \centering
    \resizebox{\textwidth}{!}{
        \begin{tikzpicture}[
    node distance=1.5cm and 1.5cm,
    font=\sffamily\small,
    >={Latex[width=2mm,length=2mm]},
    line width=0.8pt,
    % --- Styles ---
    block/.style={
        draw=darkblue, 
        line width=1pt, 
        rounded corners=3pt, 
        minimum height=1.2cm, 
        minimum width=2.6cm, 
        align=center,
        fill=white
    },
    lstm/.style={
        block, 
        fill=gray!10, 
        dashed 
    },
    encoder/.style={
        block, 
        fill=encoderblue
    },
    decoder/.style={
        block, 
        fill=decoderorange
    },
    io/.style={
        circle, 
        draw=darkblue, 
        fill=white, 
        inner sep=0pt, 
        minimum size=0.9cm, 
        line width=1pt,
        font=\bfseries
    }
]
    % --- Color Definitions (Local) ---
    \definecolor{darkblue}{RGB}{20,40,90}
    \definecolor{encoderblue}{RGB}{225,235,250}
    \definecolor{decoderorange}{RGB}{250,242,225}

    % --- Nodes ---

    % 1. LSTM (Top Center - Anchor)
    \node[lstm] (lstm) {\textbf{LSTM}\\(Frozen Teacher)};
    
    % 2. Outputs (Teacher)
    \node[io, right=1.5cm of lstm, fill=gray!5] (y_teacher) {$y_{\text{\tiny LSTM}}$};
    \node[above=0.1cm of y_teacher, font=\scriptsize, color=gray] {Teacher Prediction};

    % 3. Inputs (Left) - Symmetric Positioning
    \coordinate (input_center) at ($(lstm.west) + (-2.8cm, 0)$);
    
    \node[io] (xd) at ($(input_center) + (0, 0.5cm)$) {$X_d$};
    \node[io] (xs) at ($(input_center) + (0, -0.5cm)$) {$X_s$};

    % 4. Student Pipeline (Centered below LSTM)
    \node[encoder, below=2.0cm of lstm] (enc)  {\textbf{Concept}\\ \textbf{Encoder} ($\mathcal{E}_\phi$)};
    
    \node[io, right=1.0cm of enc, fill=encoderblue!30, rectangle, rounded corners, minimum height=0.8cm, minimum width=1.2cm] (concepts) {$p$};
    \node[above=0.1cm of concepts, font=\scriptsize, color=darkblue] {Concept Probabilities};

    \node[decoder, right=1.0cm of concepts] (dec) {\textbf{Context-Aware}\\ \textbf{Decoder} ($\mathcal{D}_\psi$)};

    \node[io, right=1.2cm of dec, fill=decoderorange!30] (y_student) {$\hat{y}$};
    \node[above=0.1cm of y_student, font=\scriptsize, color=darkblue] (y_label) {Surrogate Prediction};

    % --- Connections ---

    % Inputs to LSTM
    \coordinate (xd_target) at ([yshift=0.25cm]lstm.west);
    \coordinate (xd_turn) at ($(xd) + (1.2,0)$); 
    \coordinate (xd_corner) at (xd_turn |- xd_target);
    
    \draw[->] (xd) -- (xd_turn) -- (xd_corner) -- node[above, font=\scriptsize, midway] {$x_{d, 1:T}$} (xd_target);
    
    \coordinate (xs_target) at ([yshift=-0.25cm]lstm.west);
    \coordinate (xs_turn) at ($(xs) + (1.2,0)$);
    \coordinate (xs_corner) at (xs_turn |- xs_target);

    \draw[->] (xs) -- (xs_turn) -- (xs_corner) -- (xs_target);

    % LSTM -> y
    \draw[->] (lstm) -- (y_teacher);
    
    % LSTM -> Encoder (h) 
    \draw[->, color=darkblue, line width=1.2pt] (lstm.south) -- node[right, font=\footnotesize, align=left] {\textbf{$h$} \scriptsize(Hidden State)} (enc.north);

    % Encoder -> C -> Decoder -> y'
    \draw[->] (enc) -- (concepts);
    \draw[->] (concepts) -- (dec);
    \draw[->] (dec) -- (y_student);

    % --- Dashed Context Line (Xs -> Decoder) ---
    \path (enc.south) ++(0,-0.6cm) coordinate (below_enc_level);
    
    \draw[->, dashed, color=darkblue!80, rounded corners=5pt] 
        (xs.south) 
        -- (xs.south |- below_enc_level)       
        -- (dec.south |- below_enc_level)      
        node[midway, fill=blue!5!white, text=darkblue, font=\footnotesize, inner sep=2pt, yshift=0.5mm] {Context Attributes} 
        -- (dec.south);                        

    % --- Background Box ---
    \begin{scope}[on background layer]
        \coordinate (box_right_edge) at (y_label.east |- y_student.center);
        \node[fit=(enc)(concepts)(dec)(y_student)(below_enc_level)(box_right_edge), 
              draw=darkblue!30, dashed, rounded corners=10pt, inner sep=12pt,
              fill=blue!5!white] (student_bg) {};
        \node[anchor=south east, font=\bfseries\color{darkblue!60}] at (student_bg.south east) {Student Surrogate};
    \end{scope}

\end{tikzpicture}
    }
    \caption{\textbf{Context-Aware Concept Distillation Architecture.} The inputs are split into dynamic forcings ($X_d$) and static attributes ($X_s$). The \textcolor{blue}{Stage 1} Encoder distills the LSTM hidden state ($h$) into interpretable concept probabilities ($p$). The \textcolor{orange}{Stage 2} \textbf{Context-Aware Decoder} ($\mathcal{D}_\psi$) uses the static context ($X_s$) to modulate the mapping from concepts ($p$) to streamflow ($\hat{y}$), allowing for basin-specific adaptations.}
    \label{fig:model_architecture}
\end{figure*}
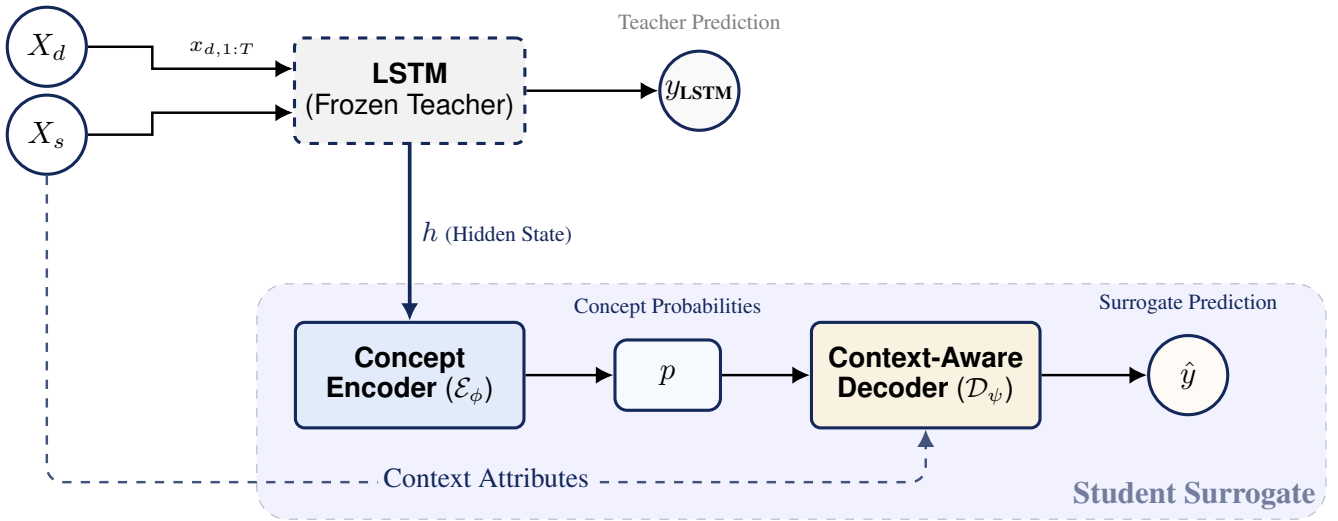
% --------------------------

The framework operates in three stages (Figure~\ref{fig:model_architecture}): (1) Unsupervised discovery of a ``Hydrological Language'' via high-fidelity feature engineering and hierarchical clustering; (2) Distilling the LSTM's internal representations into these concepts via a Concept Encoder; and (3) Reconstructing streamflow predictions via a Context-Aware Decoder that utilizes a residual hypernetwork to enforce physical consistency.

\subsection{Unsupervised Concept Discovery}
\label{sec:discovery}

To characterize the model's behavior, we utilize SHAP \cite{lundberg2017unified}, which decomposes a prediction into additive contributions from each input. For each sample, this generates an attribution matrix $\Phi \in \mathbb{R}^{T \times F}$, where $T=365$ days and $F$ represents the set of dynamic and static input features.

We elect to perform clustering in this explanation space rather than in the raw input space. This distinction is fundamental: two basins may experience identical heavy rainfall (similar inputs), yet the LSTM may predict a flood for one (impermeable urban soil) and a negligible flow response for another (permeable sandy soil). By clustering SHAP values, we group events based on the model's internal reasoning mechanism, the specific triggers it utilized to reach a decision, rather than merely correlating weather patterns.
However, the raw attribution matrix $\Phi$ is high-dimensional and sparse. To extract meaningful semantic patterns from this dense signal, we apply a specialized temporal abstraction.

\subsubsection{Log-Temporal Windowing}
The raw attribution matrix $\Phi$ is extremely high-dimensional\updated{; specifically, the dynamic subset alone contains over 5,000 values per sample ($365 \text{ days} \times 14 \text{ dynamic features}$)}. Directly clustering this sparse time series is computationally intractable. To distill the relevant signals, we must aggregate the temporal dimension for the dynamic features.

Hydrological systems exhibit a fading memory: model sensitivity is acute for recent events but diffuse for distant ones. We replace uniform aggregation with a log-temporal windowing scheme $\mathcal{W} = \{W_0, \dots, W_6\}$. The windows cover days $[0\text{-}2], [3\text{-}7],[8\text{-}14],[15\text{-}30], [31\text{-}60],[61\text{-}180],$ and $[181\text{-}364]$. \updated{These bins approximate logarithmic spacing to capture hydrologic memory decay: narrow early bins isolate acute responses (e.g., 0--2 days for surface runoff, 3--7 days for interflow), while wider bins track slow long-term dynamics (e.g., 15--30 days for soil moisture, 181--364 days for deep groundwater).} For the 51 static basin attributes, temporal windowing is unnecessary; their SHAP values are appended directly to the feature vector.

\subsubsection{Robust Statistical Features}
Naïve aggregation strategies reduce a temporal window to a single sum, discarding the internal structure of the model's reasoning. To preserve the signal's morphology, we instead extract four robust statistics for each feature $f$ and window $W$ (where $\phi_{f,t}$ denotes the SHAP attribution of feature $f$ at time step $t$):
\textbf{Net Influence ($I$):} The signed sum of SHAP values $\sum_{t \in W} \phi_{f,t}$, capturing the net directional push (positive or negative) on the prediction.
\textbf{Magnitude ($M$):} The sum of absolute SHAP values $\sum_{t \in W} |\phi_{f,t}|$, measuring the total attention paid to the feature regardless of direction.
\textbf{Soft Peak Lag ($\tilde{t}$):} \updated{A differentiable temporal center-of-mass indicating effectively the weighted timing of the signal within the window, making this statistic stable against noise.}
\begin{equation}
    \tilde{t}_{f,W} = \frac{\sum_{t \in W} t \cdot |\phi_{f,t}|^2}{\sum_{t \in W} |\phi_{f,t}|^2}
\end{equation}
\textbf{Concentration ($C^*$):} To distinguish between sustained pressures (e.g., Snowmelt) and acute shocks (e.g., Flash Floods), we compute a rescaled Herfindahl-Hirschman Index (HHI). Let \updated{$p_{f,t,W} = |\phi_{f,t}| / M_{f,W}$} be the proportional importance of time step $t$.
\begin{equation}
    C^*_{f,W} = \frac{\sum_{t \in W} \updated{p_{f,t,W}^2} - 1/|W|}{1 - 1/|W|}
\end{equation}
$C^*$ ranges from 0 (uniform influence) to 1 (single-day spike), providing a scale-invariant measure of temporal sparsity.

These statistics are concatenated to form the final feature vector $v \in \mathbb{R}^{D}$ used for clustering.

\subsubsection{Hierarchical Concept Clustering}
We define the concept vocabulary $\mathcal{C}$ via unsupervised clustering of the feature vectors $v$ (which comprise the concatenated statistics from all windows).
To determine the optimal number of concepts, we analyzed the manifold structure using Principal Component Analysis (PCA). The analysis revealed that the first principal component captures a disproportionately large share of the variance, creating a strict bipartite separation between memory-driven (baseflow) and event-driven (flash flood) dynamics. A flat clustering approach fails to capture subtler variations within these regimes.
Therefore, we employ a hierarchical K-Means strategy: we first partition the space into $K_{super}=2$ global regimes, and subsequently cluster each regime into $K_{sub}=3$ sub-patterns based on Silhouette optimization. This yields a total of $K=6$ distinct semantic concepts (e.g., ``Delayed Snowmelt,'' ``Convective Flash Flood'') that serve as the ground truth labels for our distillation process (see Figure~\ref{fig:concepts}).

\subsection{Context-Aware Concept Distillation}

\textbf{Problem Setting:} Let $f_\theta$ be a pre-trained hydrological model (specifically an LSTM) that maps dynamic forcings $x_{d}$ and static basin attributes $x_{s}$ to streamflow predictions $y_{LSTM}$. We consider this model as a fixed ``Teacher'' and aim to train an interpretable ``Student'' surrogate.

We construct a distillation dataset of extreme events $\mathcal{D} = \{(h_i, y_{LSTM,i}, x_{s,i})\}_{i=1}^N$, where each sample $i$ represents a specific target date \updated{for a specific basin}. We focus exclusively on high-flow events (highest 10\% streamflow) \updated{because they are the most societally impactful events, ensuring} the interpretation targets critical hydrological hazards. Here, $h_i$ is the LSTM's final hidden state, $x_{s,i}$ denotes the static basin attributes, and \textbf{$y_{LSTM,i}$} is the teacher's prediction.

\subsubsection{Stage 1: Distillation via Concept Encoder}
The Concept Encoder $\mathcal{E}_\phi$ acts as a translator, mapping the high-dimensional LSTM's hidden state $h_i$ to the probability distribution over the $K=6$ concepts (see Figure~\ref{fig:concepts}). We parameterize $\mathcal{E}_\phi$ as an MLP with two hidden layers ($256 \to 64 \to 32 \to K$) using ReLU activations and dropout ($p=0.2$), trained to minimize Cross-Entropy loss against SHAP cluster labels.

\begin{equation}
    p_i = \text{Softmax}(\text{MLP}_{enc}(h_i)) \in \mathbb{R}^K
\end{equation}
Here, $p_{i,k}$ represents the probability that extreme event $i$ is driven by hydrological concept $k$. By freezing this encoder after Stage 1, we ensure that the intermediate representation remains grounded in the discovered SHAP semantics.

\subsubsection{Stage 2: Reconstruction via Context-Aware Decoder}
\begin{figure}[t]
    \centering
    \resizebox{\columnwidth}{!}{
        \begin{tikzpicture}[
    node distance=1.5cm and 1.5cm,
    font=\sffamily\small,
    >={Latex[width=2mm,length=2mm]},
    line width=0.8pt,
    % --- Styles ---
    block/.style={
        draw=darkblue, 
        line width=1pt, 
        rounded corners=3pt, 
        minimum height=1.0cm, 
        minimum width=2.0cm, 
        align=center,
        fill=white
    },
    op/.style={
        circle,
        draw=darkblue,
        line width=1pt,
        minimum size=0.8cm,
        inner sep=0pt,
        fill=white,
        font=\large\bfseries
    },
    param/.style={
        rectangle,
        draw=darkblue!60,
        dashed,
        fill=gray!5,
        minimum height=0.8cm,
        minimum width=1.2cm,
        align=center,
        font=\footnotesize
    },
    io/.style={
        circle, 
        draw=darkblue, 
        fill=white, 
        inner sep=0pt, 
        minimum size=0.9cm, 
        line width=1pt,
        font=\bfseries
    }
]
    % --- Color Definitions ---
    \definecolor{darkblue}{RGB}{20,40,90}
    \definecolor{decoderorange}{RGB}{250,242,225}
    \definecolor{highlight}{RGB}{255,140,0}
    \definecolor{encoderblue}{RGB}{225,235,250} 

    % --- 1. Main Data Flow (Top Row) ---
    
    % Input Concepts (p)
    \node[io, fill=encoderblue!30] (p) {$p$};
    \node[above=0.1cm of p, font=\scriptsize, color=darkblue, align=center] {Concept\\Probabilities};

    % Multiplication Op
    \node[op, right=2.8cm of p] (mult) {$\times$};
    
    % Addition Op
    \node[op, right=2.0cm of mult] (add) {$+$};

    % Output Streamflow (\hat{y})
    \node[io, right=2.8cm of add, fill=decoderorange!50] (y) {$\hat{y}$};
    \node[above=0.1cm of y, font=\scriptsize, color=darkblue] {Prediction};

    % Connect Main Flow
    \draw[->] (p) -- (mult);
    \draw[->] (mult) -- (add);
    \draw[->] (add) -- (y);

    % --- 2. Hypernetwork Machinery (Bottom Area) ---

    % Static Input (Bottom Center)
    \node[io, below=3.5cm of mult] (xs) at ($(mult)!0.5!(add) + (0,-3.5)$) {$X_s$};
    
    % Static Processor (MLP)
    \node[block, above=0.8cm of xs, fill=white] (mlp) {\textbf{Static Processor}\\(MLP)};
    
    % Embedding z
    \node[draw=darkblue, circle, inner sep=2pt, fill=white, above=0.6cm of mlp] (z) {\footnotesize $z$};
    
    % --- 3. Parameter Generation Heads ---

    % Left Branch: Weights (W)
    \node[op, below=1.2cm of mult, font=\bfseries] (sum_w) {$+$};
    \node[param, left=0.6cm of sum_w] (w_base) {$W_{base}$\\\scriptsize(Global)};
    \node[block, below=0.5cm of sum_w, minimum width=1.8cm, minimum height=0.8cm, font=\footnotesize] (head_w) {Linear\\Head};
    
    % Right Branch: Bias (b)
    \node[op, below=1.2cm of add, font=\bfseries] (sum_b) {$+$};
    \node[param, right=0.6cm of sum_b] (b_base) {$b_{base}$\\\scriptsize(Global)};
    \node[block, below=0.5cm of sum_b, minimum width=1.8cm, minimum height=0.8cm, font=\footnotesize] (head_b) {Linear\\Head};

    % --- Connections for Hypernet ---

    % Xs -> MLP -> z
    \draw[->] (xs) -- (mlp);
    \draw[->] (mlp) -- (z);

    % z -> Heads (Rounded corners)
    \draw[->, rounded corners=5pt] (z) -| (head_w);
    \draw[->, rounded corners=5pt] (z) -| (head_b);

    % Heads -> Residuals -> Sums
    \draw[->] (head_w) -- node[left, font=\tiny, pos=0.25, inner sep=1pt] {$\times \beta$} node[right, font=\tiny, color=highlight, pos=0.8] {$\Delta W$} (sum_w);
    
    \draw[->] (head_b) -- node[right, font=\tiny, pos=0.25, inner sep=1pt] {$\times \alpha$} node[left, font=\tiny, color=highlight, pos=0.8] {$\Delta b$} (sum_b);

    % Bases -> Sums
    \draw[->] (w_base) -- (sum_w);
    \draw[->] (b_base) -- (sum_b);

    % Sums -> Main Operations
    \draw[->, color=darkblue, line width=1.2pt] (sum_w) -- node[left, font=\bfseries\footnotesize, align=right] {$W(x_s)$} (mult);
    \draw[->, color=darkblue, line width=1.2pt] (sum_b) -- node[right, font=\bfseries\footnotesize] {$b(x_s)$} (add);

    % --- Background Box ---
    \begin{scope}[on background layer]
        % We strictly include ONLY the internal modules.
        \node[fit=(mult)(add)(mlp)(w_base)(b_base)(head_w)(head_b), 
              draw=decoderorange!80!black, dashed, rounded corners=10pt, line width=1pt,
              inner sep=18pt, 
              fill=decoderorange!40] (decoder_bg) {};
        
        % Title
        \node[anchor=north, font=\bfseries\color{darkblue!80}, yshift=-5pt] at (decoder_bg.north) {Context-Aware Decoder};
        
        % Label for Residual Hypernetwork
        \node[anchor=south west, font=\bfseries\scriptsize\color{highlight}, inner sep=5pt] at (decoder_bg.south west) {Residual Hypernetwork};
    \end{scope}

\end{tikzpicture}
    }
    \caption{\textbf{Zoom-in of the Stage 2 Context-Aware Decoder.} The Residual Hypernetwork processes static attributes ($X_s$) to generate a compact latent context embedding ($z$). This embedding drives two linear heads to produce residual corrections ($\Delta W, \Delta b$), which shift the global hydrological laws ($W_{base}, b_{base}$) to create basin-specific weights ($W(x_s), b(x_s)$).}
    \label{fig:decoder_zoom}
\end{figure}
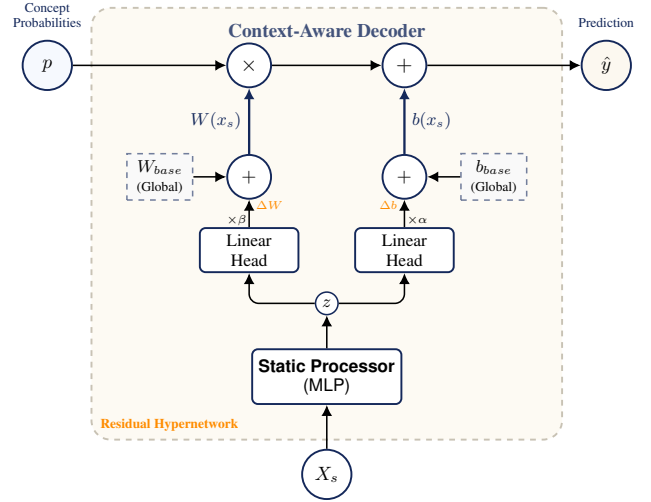

In Stage 2, we map the concept probabilities $p_i$ to streamflow $\hat{y}_i$. \updated{To ensure interpretability, this mapping must remain linear ($y = W \cdot p + b$). However,} a standard linear decoder is insufficient because the hydrological impact of a concept varies by geography (e.g., a "Snowmelt" concept may yield massive flow in steep rocky basins but low flow in flat permeable ones). A single global weight vector fails to capture these spatially dependent variations. We introduce a Context-Aware Decoder governed by a Residual Hypernetwork (Figure~\ref{fig:decoder_zoom})\updated{, which dynamically generates basin-specific weights $W(x_s)$. This preserves linear interpretability for dynamic concepts while enabling non-linear spatial modulation.} This design imposes a structural inductive bias mirroring the physical principle: \\
\begin{equation*}
\resizebox{0.98\columnwidth}{!}{$\displaystyle \text{Streamflow} \propto \underbrace{\text{Basin Sensitivity } (W)}_{\text{Spatial } (x_s)} \times \underbrace{\text{Dynamic Forcing } (p_i)}_{\text{Temporal } (p_i)}$}
\end{equation*}

% --- FIGURE 2 ---
% --------------------------

\noindent First, a Static Processor compresses the high-dimensional basin attributes $x_{s,i}$ into a compact latent context embedding $z_i$. We parameterize this as a 2-layer MLP ($51 \to 64 \to 64$) with ReLU activations and Dropout ($p=0.1$):
\begin{equation}
    z_i = \text{MLP}_{stat}(x_{s,i}) \in \mathbb{R}^{64}
\end{equation}
This embedding $z_i$ serves as a low-dimensional signature of the basin's physical identity (e.g., "Steep \& Impermeable").
Next, the hypernetwork generates basin-specific parameters via two parallel linear heads. We define the effective Bias $b(x_{s,i})$ and Weight Matrix $W(x_{s,i})$ as residual corrections to the global average:
\begin{align}
    b(x_{s,i}) &= b_{base} + \alpha \cdot \text{Linear}_{bias}(z_i) \\
    W(x_{s,i}) &= \underbrace{W_{base}}_{\text{Global Law}} + \underbrace{\beta \cdot \text{Linear}_{weight}(z_i)}_{\text{Local Adjustment}}
\end{align}
Here, $W_{base}$ represents the mean hydrological response across all basins (e.g., precipitation generally increases flow), while the second term captures local deviations driven by specific traits $z_i$ (e.g., steep slopes amplify the response). $\alpha$ and $\beta$ are learnable scalars that scale these context-aware adjustments.
The final prediction for event $i$ is the dot product of the dynamic concepts and the context-aware weights:
\begin{equation}
    \hat{y}_i = W(x_{s,i}) \cdot p_i + b(x_{s,i})
\end{equation}

\textbf{Training Objective:} The decoder minimizes the MSE between the surrogate prediction $\hat{y}_i$ and the teacher LSTM's prediction $y_{LSTM, i}$, rather than ground truth observations. This ensures strict fidelity to the LSTM's internal reasoning.

Crucially, while the hypernetwork is non-linear in $x_s$, the final prediction remains linear in concepts $p_i$. This enables direct interpretation of concept importance via the effective weights $W(x_{s,i})$, enforcing a strict separation of concerns between temporal dynamics and spatial heterogeneity.

% ----------------------------------------------------------------------------------------------------------------

\section{Experimental Setup}

\subsection{Dataset: Caravan}
We evaluate our framework on the Caravan dataset \cite{kratzert2023caravan}, utilizing a curated set of 5,203 basins spanning diverse hydro-climatic zones globally. \updated{Caravan was selected as it is the standard state-of-the-art benchmark for global hydrological machine learning, providing the only multi-continental dataset in a uniform format.} 
To ensure robust generalization to unseen future climates, we employ a temporal split: Training (2001--2014), Validation (2015--2017), and Testing (2018--2020).

\noindent\textbf{Input Features:} The model utilizes two types of inputs from the Caravan dataset. (1) \textbf{Dynamic Forcings ($x_d$):} 14 time-varying meteorological and state variables, including precipitation, temperature (mean/min/max), shortwave/longwave radiation, dewpoint, soil moisture (4 layers), snow water equivalent, surface pressure, wind components (u/v), and potential evaporation. (2) \textbf{Static Attributes ($x_s$):} 51 time-invariant basin properties describing topography (e.g., area, elevation, slope), climate indices (e.g., aridity, seasonality, precipitation frequency/duration), and soil/land characteristics (e.g., groundwater depth, 15 vegetation classes, and 21 land cover classes).

\subsection{The Teacher Model (LSTM)}
Our "Teacher" model is a standard Long Short-Term Memory (LSTM) network. It processes the sequence of dynamic inputs $x_{d, 1:T}$ ($T=365$) to update a hidden state $h_t \in \mathbb{R}^{256}$. To incorporate spatial context, the static attributes $x_s$ are passed through a fully-connected embedding network (two layers: $64 \to 32$ neurons, Tanh activation) and concatenated with the dynamic inputs at each time step.

\textbf{Training Objective:} The LSTM is trained to minimize the Basin-Averaged Nash-Sutcliffe Efficiency ($\text{NSE}^*$) \cite{kratzert2019benchmarking}. Unlike standard MSE, $\text{NSE}^*$ normalizes errors by the specific basin's variance ($\sigma_b^2$), preventing high-discharge basins from dominating the gradient:
\begin{equation}
    \mathcal{L}_{NSE^*} = \frac{1}{B} \sum_{b=1}^{B} \sum_{t=1}^{N} \frac{(y_{LSTM, b,t} - y_{obs, b,t})^2}{(\sigma_b + \epsilon)^2}
\end{equation}
where $y_{LSTM}$ is the model prediction, $y_{obs}$ is the observed streamflow, and $\epsilon$ is a smoothing term. The model achieves robust \updated{accuracy against ground truth observations} (Median NSE $\approx 0.60$), providing a high-quality distillation target.

\subsection{Defining Extreme Events}
Since our goal is to interpret critical hydrological hazards, we focus on \textbf{extreme events}. We construct the distillation dataset $\mathcal{D}$ by selecting the top 10\% of days by streamflow volume for each basin from the training period. This yields a large-scale dataset of high-impact events, ensuring the concepts discovered (e.g., Flash Flood) represent significant hydrological mechanisms rather than background noise.

\subsection{Evaluation Metrics}
We evaluate the fidelity of the surrogate model using three complementary metrics, each targeting a specific dimension of model quality: 
(1) \textbf{Global NSE} measures the ability to capture large-scale event magnitudes (volume-weighted); 
(2) \textbf{Median Local NSE} measures the spatial consistency and reliability across diverse geographies (unweighted); and
(3) \textbf{Success Rate} (percentage of basins with NSE $> 0.7$) quantifies utility, corresponding to "Good" to "Very Good" performance in hydrological standards \cite{moriasi2007model}.

\textbf{Target Definition:} To measure distillation fidelity, all metrics use the LSTM's predictions ($y_{LSTM}$) as the target rather than observed streamflow. \updated{Thus, a high fidelity score (e.g., Median NSE 0.70) demonstrates the surrogate successfully mimics the LSTM, independent of the LSTM's baseline accuracy against Nature (0.60).}

\subsection{Baselines}
To quantify the value of our architectural choices, we compare the Context-Aware Decoder against three ablative baselines:
\textbf{Naive Linear (Global Weights):} A \updated{global linear regression} decoder that learns a single, universal weight vector $W \in \mathbb{R}^K$ \updated{and bias $b$} for all basins ($\hat{y} = W \cdot p + b$). This tests whether basin-specific modulation is actually required or if a global "average" law suffices.
\textbf{Deep MLP (Black Box):} A non-linear decoder that concatenates concept probabilities and static features into a standard multi-layer perceptron ($\hat{y} = \text{MLP}(p, x_s)$). This tests whether our interpretability constraint (enforcing linearity in concepts) comes at the cost of performance.
\textbf{Simple Summation (Feature Ablation):} To validate the necessity of our high-fidelity feature engineering, we generate alternative concepts using a standard coarse-grained approach: simple summation of SHAP values over three broad windows (Recent, Short-term, Long-term). We then retrain the full surrogate model (Encoder and Decoder) using these coarse concept labels. This isolates the contribution of the log-temporal statistics (e.g., Peak Lag) to the model's success.

% ----------------------------------------------------------------------------------------------------------------

\section{Results and Analysis}

% --- FIGURE 3: 6-GRID SHAP ---
\begin{figure*}[t]
    \centering
    \includegraphics[width=0.75\textwidth]{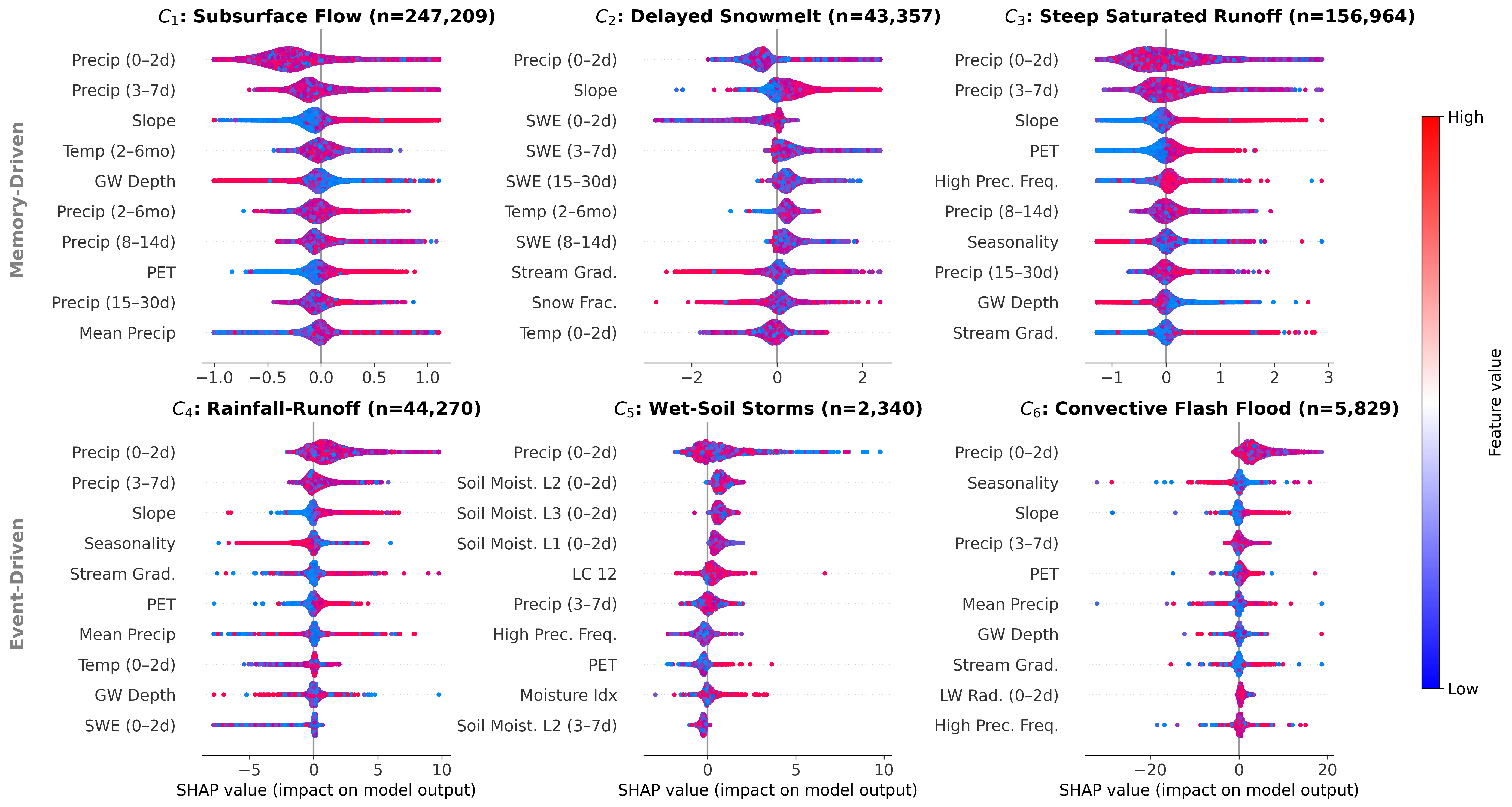}
    \caption{\textbf{The Discovered Hydrological Language.} SHAP summary plots for the $K=6$ concept clusters. These distinct patterns serve as the ground-truth targets for the Concept Encoder. The model successfully disentangles distinct physical mechanisms. Top Row (Memory-Driven): Concepts like $C_2$ (Delayed Snowmelt) are driven by temperature and snow depth with long temporal lags. Bottom Row (Event-Driven): Concepts like $C_6$ (Convective Flash Flood) are dominated by immediate precipitation (0-2 days) with high concentration. Blue/Red points indicate low/high feature values.}
    \label{fig:concepts}
\end{figure*}
% -----------------------------

We evaluate the framework along three dimensions: (1) \textbf{Fidelity}, ensuring the surrogate model accurately mimics the teacher; (2) \textbf{Necessity}, justifying the advanced feature engineering; and (3) \textbf{Interpretability}, confirming that the discovered concepts represent distinct hydrological processes.

\subsection{Fidelity Analysis}

We compare the Context-Aware Decoder against the Naive Linear and Deep MLP baselines on the full global dataset (5,203 basins). Table~\ref{tab:fidelity} summarizes the results.

% --- CDF PLOT ---
\begin{figure}[t]
    \centering
    \includegraphics[width=\columnwidth]{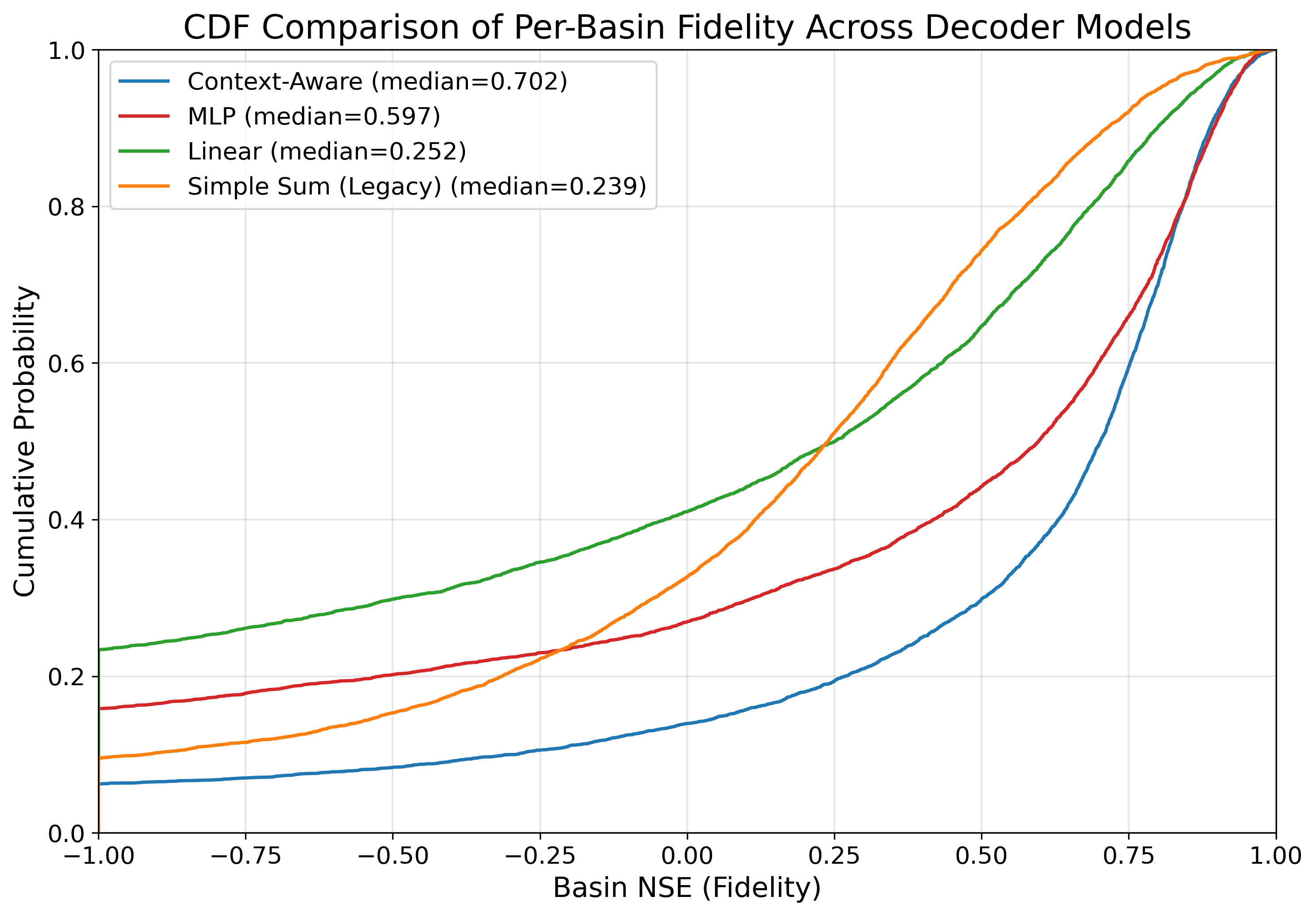} 
    \caption{\textbf{Cumulative Distribution of NSE Fidelity Scores.} Comparison of per-basin performance across 5,203 basins. The Context-Aware model (Blue) consistently dominates the black-box MLP (Red). Furthermore, the gap between Blue and Orange (Simple Sum) visually demonstrates the necessity of high-fidelity feature engineering; without it, the model fails to capture the temporal dynamics required for generalization.}
    \label{fig:fidelity_cdf}
\end{figure}
% ---------------------

\begin{table}[t]
    \centering
    \small
    \resizebox{\columnwidth}{!}{
    \begin{tabular}{l c c c}
        \toprule
        \textbf{Model Architecture} & \textbf{Global NSE} & \textbf{Median NSE} & \textbf{Success Rate} \\
        & (Volume) & (Spatial) & ($>0.7$) \\
        \midrule
        Naive Linear (Global) & 0.67 & 0.25 & 18.8\% \\
        Deep MLP (Black Box) & 0.83 & 0.60 & 39.9\% \\
        \textbf{Context-Aware (Ours)} & \textbf{0.88} & \textbf{0.70} & \textbf{50.3\%} \\
        \bottomrule
    \end{tabular}
    }
    \caption{Fidelity comparison on the full global dataset. The Context-Aware model significantly outperforms the Deep MLP in spatial generalization (Median NSE 0.70 vs 0.60). Success Rate denotes the percentage of basins achieving high fidelity (NSE $> 0.7$), corresponding to the "Good" to "Very Good" range in hydrological standards \protect\cite{moriasi2007model}.}
    \label{tab:fidelity}
\end{table}

\textbf{Generalization beyond the Black Box:} Remarkably, our interpretable surrogate model outperforms the black-box Deep MLP (Median NSE $0.70$ vs $0.60$). As shown in Figure~\ref{fig:fidelity_cdf}, the Context-Aware model exhibits stochastic dominance over the baselines, maintaining superior performance across the basin population. This suggests that the Residual Hypernetwork design acts as a regularizer. While the Deep MLP overfits to training correlations, our architecture enforces a physical separation between global laws ($W_{base}$) and local adaptations ($\Delta W$), enabling superior generalization to the unseen test period (2018--2020).

\textbf{Error Analysis (The Variance Paradox):} We investigated the tail of basins with low fidelity scores (NSE $< 0$) and found they are heavily concentrated in low-variance catchments ($\log_{10}(\sigma^2) < -1$). Paradoxically, these "failed" basins exhibit significantly lower absolute errors (Median RMSE = 0.14) compared to the high-fidelity basins (Median RMSE = 0.51). This empirical evidence confirms that the low NSE scores are not due to predictive model failure, but rather due to the inherent sensitivity of the NSE metric to vanishing denominators in low-flow regimes.

\subsection{Ablation Study: Feature Engineering}
To validate the necessity of our upstream feature engineering (log-temporal windows, soft peak lag), we compare it against the "Simple Summation" baseline (Table~\ref{tab:ablation}).

\begin{table}[t]
    \centering
    \small
    \begin{tabular}{l c c}
        \toprule
        \textbf{Feature Engineering} & \textbf{Median NSE} & \textbf{Success Rate} \\
        \midrule
        Legacy (Simple Sum) & 0.24 & 10.9\% \\
        \textbf{Advanced (Log-Window)} & \textbf{0.70} & \textbf{50.3\%} \\
        \bottomrule
    \end{tabular}
    \caption{Ablation of the Feature Engineering module. Using simple window sums causes a collapse in spatial generalization.}
    \label{tab:ablation}
\end{table}

The results are stark: the Success Rate ($NSE > 0.7$) drops from $50.3\%$ to $10.9\%$ when using simple summation. As visualized in Figure~\ref{fig:fidelity_cdf}, the Legacy model (Orange) degrades significantly across the entire distribution compared to the Advanced model (Blue). While simple sums capture the total moisture volume (preserving decent Global NSE), they destroy the temporal morphology of the signal. Including Soft Peak Lag and Concentration enables the clustering algorithm to distinguish between short, intense flash floods and prolonged saturation events, a structural distinction necessary for the model to generalize spatially across diverse basins.

\subsection{Qualitative Analysis: The Hydrological Language}
Figure~\ref{fig:concepts} visualizes the $K=6$ concepts discovered via clustering. Each plot aggregates SHAP attribution patterns for that cluster. Crucially, these are not just post-hoc visualizations; they define the semantic vocabulary the Concept Encoder learns in Stage 1. The hierarchical clustering reveals two distinct reasoning regimes:

\begin{table}[t]
    \centering
    \small
    \resizebox{\columnwidth}{!}{
    \begin{tabular}{l l l}
        \toprule
        \textbf{ID} & \textbf{Interpretative Label} & \textbf{Dominant Drivers (Evidence)} \\
        \midrule
        $C_1$ & Subsurface Flow & Negative Precip Impact, Slope, Shallow Groundwater \\
        $C_2$ & Delayed Snowmelt & Snow Water Equivalent, Temperature \\
        $C_3$ & Steep Saturated Runoff & Steep Slope, Lagged Precip (2--4 wks) \\
        $C_4$ & Rainfall-Runoff & Precip (0--7 days), Slope \\
        $C_5$ & Wet-Soil Storms & Precip + Soil Moisture (Layers 1--3) \\
        $C_6$ & Convective Flash Flood & Immediate Precip ($<2$d), Steep Slope \\
        \bottomrule
    \end{tabular}
    }
    \caption{The Hydrological Language. Cluster IDs are mapped to \textit{candidate} labels based on their dominant feature drivers (visualized in Figure~\ref{fig:concepts}). Future work will refine these semantics with our domain expert.}
    \label{tab:concept_definitions}
\end{table}

\noindent \textbf{Memory-Driven Concepts ($C_1$--$C_3$):} These regimes are dominated by state variables. For example, $C_2$ (labeled as Delayed Snowmelt) is driven by temperature and snow depth with long temporal lags. Notably, $C_1$ (Subsurface Flow) exhibits a pattern where recent precipitation is negatively correlated with the concept's activation. This reflects a dry regime where flow is sustained by slow release rather than immediate runoff.
\textbf{Event-Driven Concepts ($C_4$--$C_6$):} These regimes are dominated by meteorological forcing. $C_6$ (Convective Flash Flood) is the purest example, driven almost entirely by immediate precipitation (0--2 days) with high concentration.

It is important to emphasize that while these concepts align with hydrological intuition, they are fundamentally data-driven abstractions. As shown in Table~\ref{tab:concept_definitions}, we assign \textit{candidate} semantic labels to facilitate communication, but these definitions serve as a starting point for domain investigation rather than final physical ground truths. \updated{Crucially, the domain expert verified these candidate labels based on samples of the actual data points within each cluster, rather than solely relying on the SHAP plots.} The primary validity of this language is functional: the high fidelity achieved by the surrogate (Median NSE 0.70) confirms that this compact vocabulary is sufficient to reconstruct the teacher's reasoning, validating the concepts' utility for flood prediction.

% ----------------------------------------------------------------------------------------------------------------

\section{Conclusion}

We presented Context-Aware Concept Distillation, a framework that transforms opaque deep sequence models into transparent, hydrology-aware surrogate models. By combining high-fidelity unsupervised concept discovery with a Residual Hypernetwork, we successfully disentangled the temporal dynamics of hydrological data from the spatial heterogeneity of diverse basins.

Our extensive evaluation on the Caravan dataset yields a compelling finding: the interpretable surrogate model generalizes to future test data better than a standard black-box baseline (Median NSE 0.70 vs 0.60). This suggests that the structural constraints of our architecture—forcing the model to reason via global laws modulated by local context—prevent the overfitting common in pure black-box approaches. Furthermore, our analysis of low-fidelity basins revealed that many apparent failures are artifacts of the NSE metric in low-variance regimes, rather than predictive shortcomings.

\textbf{Real-World Impact:} CACD enables trustworthy flood prediction models whose behavior can be expressed through a compact set of automatically discovered, human-interpretable concepts. \updated{Importantly, XAI should expose model flaws, not mask them. To avoid "interpretable misinformation", where a systematic LSTM error receives a plausible explanation, these narratives must be evaluated alongside historical error bounds.} By demonstrating that accurate streamflow predictions can be reconstructed using only these concepts, the framework provides empirical evidence that model decisions rely on stable and inspectable abstractions rather than opaque internal dynamics. This reduces the risk of "right-for-the-wrong-reasons" behavior, directly increasing trust, auditability, and accountability for AI-assisted flood forecasting in public safety contexts. This capability allows emergency responders and authorities not just to use AI predictive outputs but to rely on them, understanding and trusting the insights and reasoning behind the automated alerts. \updated{Future work includes qualitative stakeholder evaluations to further validate the operational utility of these concepts.}

\section*{Acknowledgements}
Claudia V. Goldman was supported by the David Goldman Data-driven Innovation Research Center at the Hebrew University Business School at the Hebrew University. This study was partly supported by the ISF grant no. 1999/22.

\bibliographystyle{named}
\bibliography{ijcai26}

\end{document}